\documentclass[10pt,twocolumn,letterpaper]{article}

\usepackage{iccv}
\usepackage{times}
\usepackage{epsfig}
\usepackage{graphicx}
\usepackage{amsmath}
\usepackage{amssymb}
\usepackage{multirow}
\usepackage{arydshln}
\usepackage{amsfonts}

\usepackage[pagebackref=true,breaklinks=true,letterpaper=true,colorlinks,bookmarks=false]{hyperref}

\iccvfinalcopy 

\def\iccvPaperID{****} 

\ificcvfinal\fi

\makeatletter
\def\thanks#1{\protected@xdef\@thanks{\@thanks
        \protect\footnotetext{#1}}}
\makeatother

\begin{document}

\title{Tracking without Label: Unsupervised Multiple Object Tracking via Contrastive Similarity Learning}

\author{
Sha Meng{*}, 
Dian Shao{*}, 
Jiacheng Guo, 
Shan Gao\dag \thanks{*Equal Contribution. \dag Corresponding Author} \\
Northwestern Polytechnical University, Xi'an, China\\
{\tt\small \{mengsha,gjc1\}@mail.nwpu.edu.cn, \{shaodian,gaoshan\}@nwpu.edu.cn}
}

\maketitle
\ificcvfinal\thispagestyle{empty}\fi

\begin{abstract}
Unsupervised learning is a challenging task due to the lack of labels. Multiple Object Tracking (MOT), which inevitably suffers from mutual object interference, occlusion, \textit{etc.}, 
is even more difficult without label supervision. 
In this paper, we explore the latent consistency of sample features across video frames and propose an Unsupervised Contrastive Similarity Learning method, named UCSL, including three contrast modules: self-contrast, cross-contrast, and ambiguity contrast.
Specifically, i) self-contrast uses intra-frame direct and inter-frame indirect contrast to obtain discriminative representations by maximizing self-similarity.
ii) Cross-contrast aligns cross- and continuous-frame matching results, mitigating the persistent negative effect caused by object occlusion. And
iii) ambiguity contrast matches ambiguous objects with each other to further increase the certainty of subsequent object association through an implicit manner.
On existing benchmarks, our method outperforms the existing unsupervised methods using only limited help from ReID head, and even provides higher accuracy than lots of fully supervised methods.
\end{abstract}

\begin{figure}[t]
\begin{center}
   \includegraphics[width=0.9\linewidth]{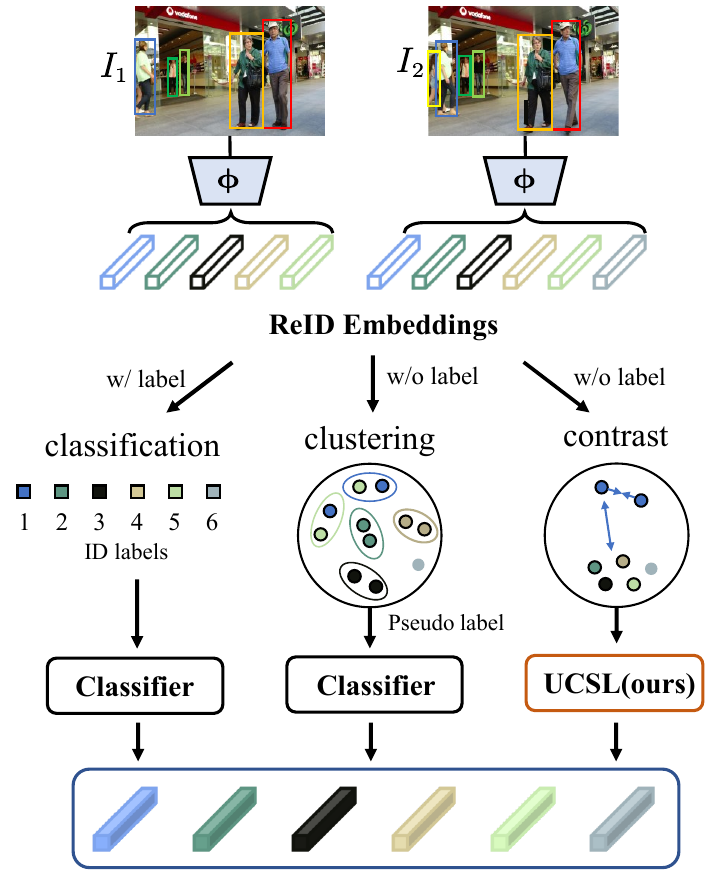}
\end{center}
   \caption{Supervised and Unsupervised MOT. In the joint detection and ReID embeddings framework, to obtain discriminative embeddings for tracking, the left branch is a usual method of supervised MOT training, \emph{i.e.}, given labels, it is trained as an object classification task. The middle branch is a common method of unsupervised training, \emph{i.e.}, it is processed by clustering, and targets with high similarity are regarded as the same class. The right branch is the proposed method 
   with contrast similarity learning to improve the similarity of the same objects without label information.}
   \label{fig:Cl}
\end{figure}

\section{Introduction}
\label{sec:intro}
As a basic task in computer vision, Multiple Object Tracking (MOT) is widely applied in a variety of fields, including robot navigation, intelligent surveillance, and other aspects \cite{wang2013intelligent,uchiyama2012object}.
Currently, one of the most popular tracking paradigms is joint detection and re-identification (ReID) embeddings. In the case of supervision, ReID is regarded as a classification task. To keep track of objects, many works \cite{wang2020towards,zhang2021fairmot} utilize appearance features for object association, where the representation ability of the ReID head will directly affect the accuracy of the object association.

However, due to limitations in various conditions such as labeled datasets, to meet the needs of researchers, there has been a growing requirement to annotate tracking datasets, which is costly and time-consuming.  Therefore, unsupervised learning of visual representation has attracted great attention in tracking. Some works \cite{wang2019unsupervised,shen2022unsupervised} have demonstrated that training the network in the real direction can also be done even without ground truth.
Some works \cite{caron2018deep,dai2021cluster} directly use ReID features to cluster objects with high similarity into the same class,
then generate pseudo-labels to train the network, as shown in the middle branch of Figure \ref{fig:Cl}. 
But these cluster-based methods are easy to accumulate errors during the training process.
In contrast, we consider using an Unsupervised Contrastive Similarity Learning (UCSL) to train the ReID branch without generating pseudo-labels, as shown in the right branch of Figure \ref{fig:Cl}.

As a video task, the objects in MOT are always changing over time, which leads to inevitable problems of mutual occlusion between objects and objects, and between objects and non-objects, as well as the disappearance of old objects and the appearance of new ones. Occluded objects are not represented consistently from frame to frame due to additional interference features. Lost and emerging objects theoretically cannot be matched with other objects, and they are almost always negative for the current association stage. 
Thus, it is difficult for the model to determine whether arbitrary two objects are the same or not.
In supervised cases, ID labels can be used to make the training more explicitly directed, while in the unsupervised case, the dual limitation of unlabeling and inherent problems makes unsupervised MOT even more challenging. 
So we manage to find potential connections between objects to determine if the objects are identical.

In this paper, we propose an Unsupervised Contrastive Similarity Learning (UCSL) method to solve the inherent object association problems of unsupervised MOT. Specifically, UCSL consists of three modules, self-contrast, cross-contrast and ambiguity contrast, designed to address different issues respectively. 
For the self-contrast, 
we first match between objects within frames and between objects in adjacent frames. Correspondingly, we get the direct and indirect matching results of the intra-frame objects. Then we maximize the matching probability of self-to-self to maximize the similarity of the same objects.
For cross-contrast, 
considering theoretically the cross-frame matching results should be consistent with the final results of continuous matching, we improve the similarity of the occluded objects by making these two matching results as close as possible.
For ambiguity contrast, 
we match between ambiguous objects, mainly containing occluded, lost, and emerging objects whose final similarity is generally low, to further determine the object identity.
Our proposed method is simple but effective, which achieves outstanding performance by utilizing only the ReID embeddings without adding any additional branch such as the occlusion handling or optical-flow based cue to the detection branch.

We implement the method on the basis of FairMOT \cite{zhang2021fairmot}  using the pre-trained model on the COCO dataset \cite{lin2014microsoft}. Our experiments on the MOT15 \cite{leal2015motchallenge}, MOT17 \cite{milan2016mot16} and MOT20 \cite{dendorfer2020mot20} datasets are conducted to evaluate the effectiveness of the proposed method. The performance of our unsupervised approach is comparable with, or outperforms, that of some supervised methods using expensive annotations.

Overall, our contributions are summarized as follows:
\begin{itemize}
    \item We propose a contrastive similarity learning method for unsupervised MOT task, which pursues latent object consistency based only on the sample features in the ReID module given without the ID information.
    \item We design three useful modules to model associations between objects in different cases. To elaborate, self-contrast module matches intra-frame objects, cross-contrast module associates cross-frame objects, and ambiguity contrast module deals with those hard/corner cases (\textit{e.g.}, occluded objects, lost objects, \textit{etc}.)
    \item Experiments on MOT15\cite{leal2015motchallenge}, MOT17\cite{milan2016mot16} and MOT20 \cite{dendorfer2020mot20} demonstrate the effectiveness of the proposed UCSL method. As an unsupervised method, UCSL outperforms state-of-the-art unsupervised MOT methods and even achieves similar performance as the fully supervised MOT methods.

\end{itemize}

\begin{figure*}[t]
\centering
    \includegraphics[width=1\linewidth]{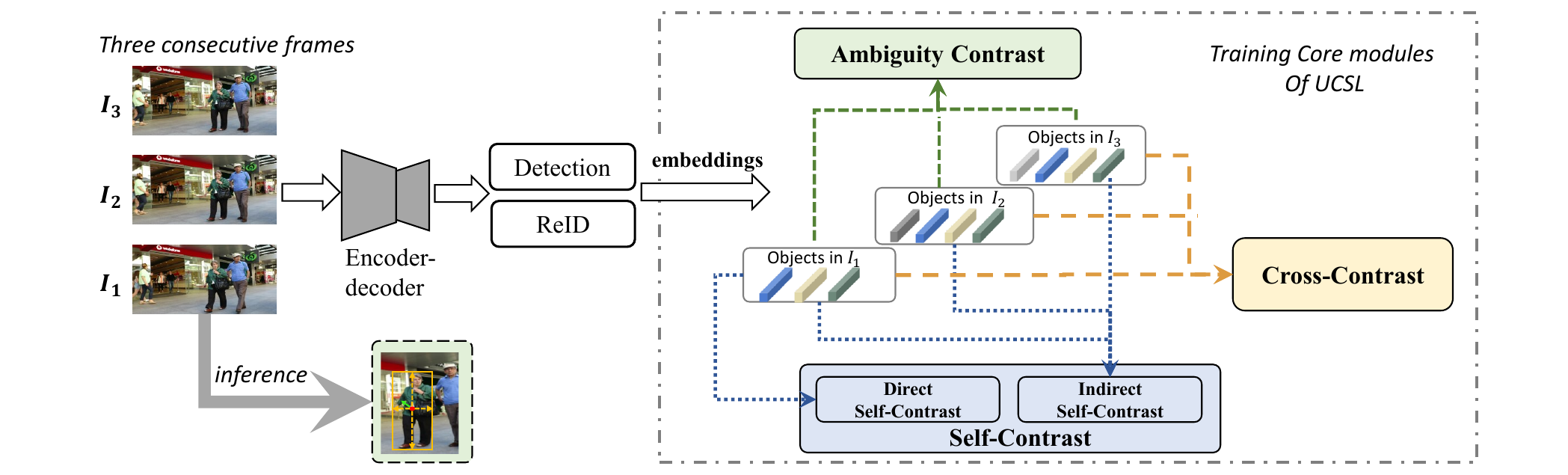}

  \caption{The overall pipeline of our proposed unsupervised contrastive similarity learning model (UCSL), which learns representations with self-contrast, cross-contrast and ambiguity contrast. }
  \label{fig:overview}
\end{figure*}
\section{Related Work}
\label{sec:formatting}
\textbf{Multi-Object Tracking.} Multi-object tracking is a task that localizes objects from consecutive frames and then associates them according to their identity. Thus, for a long time, the most classic tracking paradigm is tracking-by-detection \cite{ramanan2003finding}, \emph{i.e.}, firstly, an object detector is used to detect objects from every frame, and secondly, a tracker is used to associate these objects across frames. A large number of works \cite{bewley2016simple,wojke2017simple,sun2019deep,braso2020learning} in this paradigm have achieved decent performance, but the paradigm relies too much on the performance of detectors. In the past two years, the joint detection and tracking or embedding paradigm has become stronger. Some transformer-based MOT architectures \cite{sun2020transtrack,meinhardt2022trackformer,xu2021transcenter} designed two decoders to perform detection and object propagation respectively. JDE \cite{wang2020towards} and FairMOT \cite{zhang2021fairmot} directly incorporated the appearance model into a one-stage detector, and then the model can simultaneously output detection results and the corresponding embeddings. These simple but effective frameworks have been what we are looking for, so we take FairMOT \cite{zhang2021fairmot} as our baseline.

\textbf{Unsupervised Tracking.} For some tasks, existing datasets or other resources cannot meet the needs of researchers. In this condition, unsupervised learning has been a popular solution and its efficiency has been demonstrated in related studies \cite{karthik2020simple, liu2022online,wang2019unsupervised,shen2022unsupervised}. SimpleReID \cite{karthik2020simple} first used unlabeled videos and the corresponding detection sets, and generated tracking results using SORT \cite{bewley2016simple} to simulate the labels, and trained the ReID network to predict the labels of the given images. It is the first demonstration of the effectiveness of the simple unsupervised ReID network for MOT. Liu \textit{et al}. \cite{liu2022online} proposed a model, named OUTrack, using an unsupervised ReID learning module and a supervised occlusion estimation module together to improve tracking performance.

\textbf{Re-Identification.} 
In the field of re-identification, which is more relevant to MOT, unsupervised learning has been widely used through various means including domain adaption, clustering,  \emph{etc.} Considering the visual similarity and cycle consistency of labels, MMCL \cite{wang2020unsupervised} predicted pseudo labels and regarded each person as a class, transforming ReID into a multi-classification problem. Some other works \cite{caron2018deep,dai2021cluster,lin2019bottom} also utilized clustering algorithms to generate pseudo labels and take them as ground truth to train the network. However, error accumulation is easy to occur during the iterative process. Recent methods propose self-supervised learning, Wang \textit{et al}. \cite{wang2020cycas} proposed CycAs inspired by the data association concept in multi-object tracking. By using the self-supervised signal as a constraint on the data, networks gradually strengthen the feature expression ability during the training process.

\textbf{Cycle Consistency.} Cycle consistency is originally proposed in Generative Adversarial Network (GAN), and widely used in segmentation, tracking, \textit{etc}. 
Jabri \textit{et al}. \cite{jabri2020space} constructed a space-time graph from the video, and cast correspondence as prediction of links. By cycle consistency, the single path-level constraint implicitly supervised chains of intermediate comparisons.
 Wang \textit{et al}. \cite{wang2019learning} used cycle consistency in time as the free supervisory signal for learning visual representations from scratch. Then they used the acquired representation to find nearest neighbors across space and time in a range of visual correspondence tasks. 

\textbf{Contrastive Learning.} Contrastive learning has shown great potential in self-supervised learning. Pang \textit{et al}. \cite{pang2021quasi} proposed QDTrack, which densely sampled hundreds of region proposals on a pair of images for contrastive learning. And they directly combined this with existing detection methods.
Yu \textit{et al}. \cite{yu2022towards} proposed multi-view trajectory contrastive learning and designed a trajectory-level contrastive loss to explore the inter-frame information in the whole trajectories.
Bastani \textit{et al}. \cite{bastani2021self} proposed to construct two different inputs for the same video sequence by hiding different information. Then they computed the trajectory of that sequence by applying the RNN model independently on each input, and trained the model using contrastive learning to produce consistent tracks.

\section{Method}
\label{sec:formatting}
In this section, we first introduce the overall pipeline, as illustrated in Figure \ref{fig:overview}, and then describe the corresponding specific concepts in detail in the subsequent parts. Finally, we introduce the whole steps of training and inference.

\subsection{Contrast Similarity Learning}
\label{sec:CSL}
Given consecutive three images $\boldsymbol{I}_1, \boldsymbol{I}_2, \boldsymbol{I}_3 \in \Bbb{R}^{H \times W \times 3}$, we first feed them to the backbone, then through detection branches and ReID heads, we could get detection results and ReID feature maps, as shown in Figure \ref{fig:overview}. 
Based on the position of the bounding box in the ground truth, the feature embedding corresponding to each object is obtained from the corresponding feature map, which forms embedding matrices $\boldsymbol{X}_1=\left[\boldsymbol{x}_1^0, \boldsymbol{x}_1^1, \ldots \ldots, \boldsymbol{x}_1^{N-1}\right] \in \Bbb{R}^{D \times N}$, $\boldsymbol{X}_2=\left[\boldsymbol{x}_2^0, \boldsymbol{x}_2^1, \ldots \ldots, \boldsymbol{x}_2^{M-1}\right] \in \Bbb{R}^{D \times M}$ and $\boldsymbol{X}_3=\left[\boldsymbol{x}_3^0, \boldsymbol{x}_3^1, \ldots \ldots, \boldsymbol{x}_3^{K-1}\right] \in \Bbb{R}^{D \times K}$, 
where $N$, $M$, and $K$ are the object numbers in $\boldsymbol{I}_1$, $\boldsymbol{I}_2$ and $\boldsymbol{I}_3$ , respectively, and $D$ is the embedding dimension.

The ReID branch is connected to three contrast similarity learning branches, in which 
(1) Self-contrast uses intra-frame direct and inter-frame indirect self-matching to obtain discriminative representations and reduce feature interference from other objects by maximizing self-similarity.
(2) Cross-contrast uses cross- and continuous-frame matching, and then adjusts similarity between objects to extract more beneficial features for object association.
(3) Ambiguity contrast takes into account occluded, lost, and emerging objects simultaneously, and these ambiguous objects are matched with each other again to further increase the certainty of subsequent object association.
We will describe the specific operation in Section \ref{sec:SC}, \ref{sec:CC} and \ref{sec:AC}, respectively.

\subsubsection{Self-Contrast Module}
\label{sec:SC}

\begin{figure}[t]
\centering
    \includegraphics[width=1\linewidth]{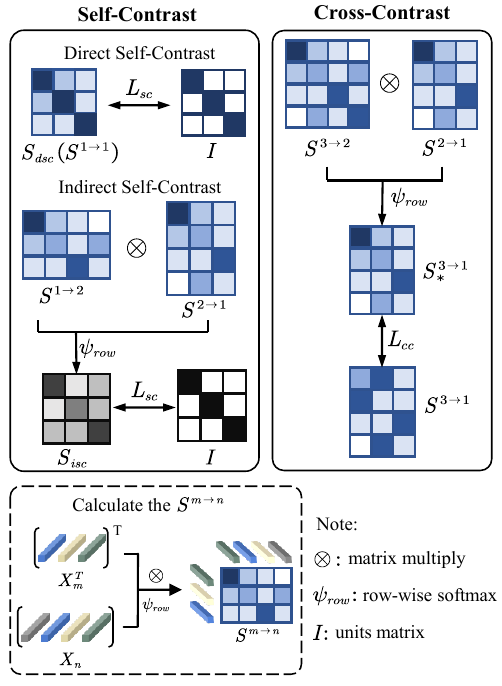}

  \caption{Self-Contrast and Cross-Contrast. We use three sets of indirect self-contrast and two sets of cross-contrast methods using different inputs. For the sake of brevity, we only show a set of specific feature calculation in each contrast.}
  \label{fig:sc and cc}
\end{figure}

According to the latent knowledge that objects from the same frame must belong to different classes, we can determine that the similarity between self-to-self should be large enough. So the proposed self-contrast finally lands on a self-to-self comparison, which is a strong, deterministic self-supervised restriction. This strong restriction allows us to improve the similarity of the same targets and reduce the interference from other objects by direct and indirect self-contrast learning, as shown in the first column of Figure \ref{fig:sc and cc}.

\textbf{Direct Self-Contrast.} 
 We use current feature matrix $\boldsymbol{X}_1=\left[\boldsymbol{x}_1^0, \boldsymbol{x}_1^1, \ldots \ldots, \boldsymbol{x}_1^{N-1}\right] \in \Bbb{R}^{D \times N}$ to directly compute the self-similarity matrix $\boldsymbol{S}_{ds}=\boldsymbol{X}_1^T \boldsymbol{X}_1 \in \Bbb{R}^{N \times N}$, where $T$ represents transpose operation.
Then we compute the assignment matrix with a softmax operation, as
\begin{equation}
\boldsymbol{S}_{dsc} =\boldsymbol{\psi}_{\text {row }}(\boldsymbol{S}_{ds}),
\label{con:softmax}
\end{equation}
where $\boldsymbol{\psi}_{\text {row }}$ is row-wise softmax operation.

\textbf{Indirect Self-Contrast.} 
MOT itself operates on multiple frames, so we further perform our self-contrast similarity learning by indirect self-to-self matching.
To measure similarity between objects, we calculate cosine similarity to get a similarity matrix between objects of different frames $\boldsymbol{S}_{is}=\boldsymbol{X}_1^T \boldsymbol{X}_2 \in \Bbb{R}^{N \times M}$.
And similar to Eq.\ref{con:softmax}, we calculate the association matrix $\boldsymbol{S}^{1 \rightarrow 2} =\boldsymbol{\psi}_{\text {row }}(\boldsymbol{S}_{is})$ and $\boldsymbol{S}^{2 \rightarrow 1} =\boldsymbol{\psi}_{\text {row }}({\boldsymbol{S}_{is}}^T)$.
The corresponding results $\boldsymbol{S}^{1 \rightarrow 2}$ and $\boldsymbol{S}^{2 \rightarrow 1}$ are considered to match the targets in $\boldsymbol{I}_1$ to $\boldsymbol{I}_2$, and the targets in $\boldsymbol{I}_2$ to $\boldsymbol{I}_1$, respectively. Each element of $\boldsymbol{S}^{1 \rightarrow 2}$ and $\boldsymbol{S}^{2 \rightarrow 1}$ in the \emph{i}-th row and \emph{j}-th column are as follows, respectively:
\begin{equation}
\begin{aligned}
&s_{i j}^{1 \rightarrow 2}=\frac{\exp \left(\left(\boldsymbol{x}_1^i\right)^T \cdot \boldsymbol{x}_2^j / \tau \right)}{\sum_{j=0}^{M-1} \exp \left(\left(\boldsymbol{x}_1^i\right)^T \cdot \boldsymbol{x}_2^j  / \tau \right)}, \\
&s_{i j}^{2 \rightarrow 1}=\frac{\exp \left(\left(\boldsymbol{x}_2^j\right)^T \cdot \boldsymbol{x}_1^i / \tau \right)}{\sum_{i=0}^{N-1} \exp \left(\left(\boldsymbol{x}_2^j\right)^T \cdot \boldsymbol{x}_1^i  / \tau \right)},
\end{aligned}
\end{equation}
where $\tau$ is a temperature hyper-parameter \cite{wang2020cycas}.

According to the cycle association consistency, after forward association $\boldsymbol{S}^{1 \rightarrow 2}$ and backward association $\boldsymbol{S}^{2 \rightarrow 1}$, each object will match itself again ideally,
\begin{equation}
\boldsymbol{S}_{isc}=\boldsymbol{S}^{1 \rightarrow 2} \boldsymbol{S}^{2 \rightarrow 1}.
\end{equation}
The corresponding self-contrast loss can be formulated as:
\begin{equation}
L_{\text {sc}}=-\frac{1}{N}\left( \sum \log \left(\operatorname{diag}\left(\boldsymbol{S}_{dsc}\right)\right)\\
+\sum \log \left(\operatorname{diag}\left(\boldsymbol{S}_{isc}\right)\right)\right),
\end{equation}
where $\operatorname{diag}()$ is to get a diagonal matrix.

Due to the self-contrast, it is obvious that the similarity between the same targets should be the largest, \emph{i.e.}, the diagonal elements of $S_{dsc}$ and $S_{isc}$ obtained above are the largest and should be as close to 1 as possible. 

\subsubsection{Cross-Contrast Module}
\label{sec:CC}
In almost all scenes of MOT, there is more or less object occlusion, and the similarity of these objects is generally low. Since MOT is an operation on multiple consecutive frames, the negative impact of these occluded objects could last for a long time. Considering theoretically the cross-frame matching results should be the same as the final results of continuous matching, we use a weaker unsupervised restriction, \emph{i.e.}, direct (cross-frame) vs. indirect (continuous-frame) association similarity comparison, to alleviate the above issue.

Specifically, we take three frames $\boldsymbol{I}_1, \boldsymbol{I}_2, \boldsymbol{I}_3 \in \Bbb{R}^{H \times W \times 3}$ as inputs, similar with Section \ref{sec:SC}, we calculate the target matching matrices between different frames, \emph{i.e.}, $\boldsymbol{S}^{1 \rightarrow 2}, \boldsymbol{S}^{2 \rightarrow 1}, \boldsymbol{S}^{2 \rightarrow 3}, \boldsymbol{S}^{3 \rightarrow 2}, \boldsymbol{S}^{1 \rightarrow 3}, \boldsymbol{S}^{3 \rightarrow 1}$. As shown in the second column of Figure \ref{fig:sc and cc}, we utilize $\boldsymbol{S}^{2 \rightarrow 1}$ and $\boldsymbol{S}^{3 \rightarrow 2}$ to compute the association matrix of $3 \rightarrow 1$, similarly use $\boldsymbol{S}^{1 \rightarrow 2}$ and $\boldsymbol{S}^{2 \rightarrow 3}$ to compute the association matrix of $1 \rightarrow 3$, as
\begin{equation}
\begin{aligned}
&\boldsymbol{S}_*^{1 \rightarrow 3}=\boldsymbol{\psi}_{\text {row }}(\boldsymbol{S}^{1 \rightarrow 2} \boldsymbol{S}^{2 \rightarrow 3}), \\
&\boldsymbol{S}_*^{3 \rightarrow 1}=\boldsymbol{\psi}_{\text {row }}(\boldsymbol{S}^{3 \rightarrow 2} \boldsymbol{S}^{2 \rightarrow 1}).
\end{aligned}
\end{equation}
These matching matrices, which are generated indirectly through a middle frame, should be the same as direct-generated matching results. 

We use relative entropy to measure the difference between the two matching distributions. KL divergence \cite{kullback1951information} is often used to compute the difference between two distributions P and Q, 
\begin{equation}
K L(P \| Q)=\sum p(x) \log \frac{p(x)}{q(x)},
\end{equation}
but it is asymmetrical. 
We further utilize JS divergence \cite{lin1991divergence} with symmetrical properties,
\begin{equation}
J S D(P \| Q)=\frac{1}{2} K L(P \| T)+\frac{1}{2} K L(Q \| T),
\end{equation}
where $T=(P + Q)/2$. 
The corresponding cross-contrast loss is as follows,
\begin{equation}
L_{cc}=\frac{1}{N} J S D(\boldsymbol{S}_*^{1 \rightarrow 3} \| \boldsymbol{S}^{1 \rightarrow 3})+\frac{1}{K} J S D(\boldsymbol{S}_*^{3 \rightarrow 1} \| \boldsymbol{S}^{3 \rightarrow 1}).
\end{equation}

By enabling the continuous and cross-frame matching results to be close together, we use the different association results to mainly mitigate the differences in the same target caused by occlusion.

\begin{figure}[t]
  \centering
    \includegraphics[width=0.8\linewidth]{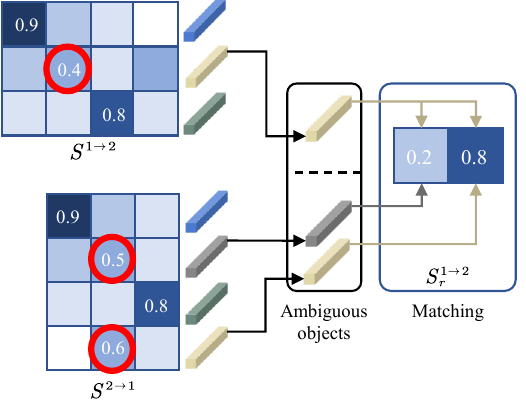}
  \caption{Ambiguity Contrast. For brevity, we only give the maximum similarity for each row, where certain objects have lower similarity to all other targets, \emph{i.e.}, even the maximum similarity is below the threshold value, which is indicated by red circles in the figure. The corresponding feature embeddings are extracted and then matched again.}
  \label{fig:rm}
\end{figure}
\subsubsection{Ambiguity Contrast}
\label{sec:AC}

There are occluded, lost, and emerging objects in the MOT, which will interfere with the whole learning process. We explore this problem and propose the ambiguity contrast module.

Based on the similarity between objects, we assume that objects with similarity greater than $\theta$ are the same object. The remaining objects with lower similarity are defined as ambiguous objects here. The low similarity mainly due to occlusion or the disappearance and appearance. In the occlusion case, objects of the same ID do exist, but the similarity is decreased due to the absence of original features and involvement of unrelated features. In the latter case, the similarity between the lost object and the newly emerged object is lower because there is really no target that can match it.

Our proposed method for ambiguous objects in the unsupervised training process is shown in Figure \ref{fig:rm}. We find the ambiguous objects in $\boldsymbol{I}_1$ according to the matching matrix of $\boldsymbol{S}^{1 \rightarrow 2}$ based on the similarity. Similarly, we get ambiguous objects in $\boldsymbol{I}_2$ based on the matching result of $\boldsymbol{S}^{2 \rightarrow 1}$. Then these objects are again subjected to similarity calculation to get the similarity matrix ${\boldsymbol{S}_r}^{1 \rightarrow 2} \in \Bbb{R}^{D \times N_r}$  and ${\boldsymbol{S}_r}^{2 \rightarrow 1} \in \Bbb{R}^{D \times M_r}$, where $N_r$ and $M_r$ are the number of ambiguous objects in $\boldsymbol{I}_1$ and $\boldsymbol{I}_2$, respectively. Finally, the loss of the module is calculated by minimum entropy:
\begin{equation}
\begin{gathered}
L_{ac}=-\frac{1}{|N_r-M_r|+1}\left(\frac{1}{N_r} {\boldsymbol{S}_r}^{1 \rightarrow 2} \log \left({\boldsymbol{S}_r}^{1 \rightarrow 2}\right)\right. \\
\left.+\frac{1}{M_r} {\boldsymbol{S}_r}^{2 \rightarrow 1} \log \left({\boldsymbol{S}_r}^{2 \rightarrow 1}\right)\right).
\end{gathered}
\end{equation}

When the number of ambiguous objects in two frames is equal, considering the two frames are very close to each other, we can assume that there are no disappearing or emerging objects, only occlusion exists, and the entropy should be as small as possible at this time. When the number of ambiguous objects in the two frames is not equal, it must contain disappearing or emerging objects. They are less similar to other objects because they cannot be matched, so we dynamically weaken the loss by adaptive coefficients.

\subsection{UCSL for Unsupervised MOT}
We apply UCSL on FairMOT \cite{zhang2021fairmot}, which composes of a backbone network, a detection head, and a re-identification head. For simplicity, the setting of the backbone and detection head follows FairMOT \cite{zhang2021fairmot}. The overall architecture of UCSL is illustrated in Figure \ref{fig:overview}.
 
In the training stage, we follow the three contrast learning sub-modules in Section \ref{sec:CSL}, and the complete loss function of ReID can be defined as follows:
\begin{equation}
L\left(I_t, I_{t-1}, I_{t-2}\right) = L_{sc}+L_{cc}+L_{ac}
\end{equation}
where three consecutive frames $I_{t }$, $I_{t-1}$ and $I_{t-2}$ denote inputs. $L_{sc}$, $L_{cc}$, and $L_{ac}$ denote self-contrast, cross-contrast, and ambiguity contrast losses mentioned above, respectively. 

In the inference stage, video frames are fed into the network one by one. Then we obtain the corresponding detection results and ReID embeddings. We use the detection bounding boxes in the first frame to initialize multiple trajectories, and then use two-stage matching to complete object association. The overall association idea is also similar to FairMOT \cite{zhang2021fairmot}, using Kalman Filter \cite{kalman1960new} to predict the position of the objects and match bounding boxes with existing trajectories using embedding distance. For the trajectories and detections that are not matched, we match them using iou distance. Finally, the remaining unmatched detections are initialized as new objects, and the unmatched trajectories are saved for 30 frames and matched when they appear again.

\section{Experiments}
\label{sec:formatting}
In this section, the proposed UCSL is evaluated on the MOT17 \cite{milan2016mot16}, MOT15 \cite{leal2015motchallenge} and MOT20 \cite{leal2015motchallenge}. The description of the datasets and the experimental setup is as follows, and next, we compare UCSL with the advanced approaches. Then, we show the evaluation of the effect of our model with ablation experiments.

\subsection{Datasets}

The proposed method is evaluated on MOT15, MOT17 and MOT20. MOT15 is the first dataset provided by MOT Challenge. It contains 22 video sequences, 11 of which are used for training and 11 for testing. The MOT15 is derived from older datasets and has different characteristics, such as fixed or moving cameras, different lighting environments,  \emph{etc.} MOT17 consists of 14 video sequences in total, 7 of which are used for training and 7 for testing, which is the most frequently used in MOT by far. MOT20 contains 4 training videos and 4 testing videos with more complex environments and greater crowd density, so MOT20 is more challenging than any previous datasets.

To evaluate our method, we use the standard MOT challenge metrics \cite{bernardin2008evaluating,li2009learning,luiten2021hota}, mainly including Multi-Object Tracking Accuracy (\textbf{MOTA}), ID F1 Score (\textbf{IDF1}), Higher Order Tracking Accuracy (\textbf{HOTA}), Mostly Tracked objects (\textbf{MT}), Mostly Lost objects (\textbf{ML}), Number of False Positives (\textbf{FP}), Number of False Negatives (\textbf{FN}) and Number of Identity Switches (\textbf{IDS}), where the higher the first four items the better, the lower the last four items the better, and we use ``$\uparrow$'' and ``$\downarrow$'' to represent respectively.

\subsection{Implementation Details}
By default, UCSL is implemented based on the basis of the original FairMOT \cite{zhang2021fairmot}. We take DLA-34 \cite{yu2018deep} as the backbone of the model and take the detection branch of the COCO dataset \cite{lin2014microsoft} pre-trained model to initialize our model parameters.  We follow the most hyper-parameters settings of FairMOT \cite{zhang2021fairmot}.

We use conventional data enhancement approaches such as rotation, random cropping and horizontal flip, scale transformation, color jittering,  \emph{etc.}, and resize the input image size to 1088×608. We use the Adam optimizer \cite{kingma2014adam} with the initial learning rate set to $10^{-4}$, and the batch size set to 8. The similarity threshold $ \theta$ in ambiguity contrast is 0.7. The model iterates 60 epochs on the MOT17 training set in the internal ablation experiments. We eventually train the corresponding dataset for 30 epochs on the basis of a pre-trained model of the CrowdHuman \cite{shao2018crowdhuman} dataset. The learning rate decays to $10^{-5}$ at the 20th epoch. Finally, we train our model on 4 RTX2080ti GPUs in about 10 hours.

\subsection{Performance and Comparison}

\begin{table}[t]
\begin{center}
    \begin{tabular}{c|c|c|c|c}
    \hline  Method & Unsup & MOTA$\uparrow$ & IDF1$\uparrow$ & HOTA$\uparrow$   \\
    \hline \multicolumn{5}{c}{MOT17} \\ \hline

  TrackFormer \cite{meinhardt2022trackformer} & No & 74.1 & 68.0 & 57.3 \\
     TransTrack \cite{sun2020transtrack} & No  & \underline{75.2} & 63.5 & 54.1 \\
   TransCenter  \cite{xu2021transcenter} & No   &	73.2 & 62.2 & 54.5 \\
    QDTrack \cite{pang2021quasi} & No & 68.7 & 66.3 & 53.9 \\
    
   JDE \cite{wang2020towards} & No  & 56.7 & 55.0 & 45.1 \\ 
     CSTrack \cite{liang2022rethinking} & No  & 74.9 & \underline{72.6} & \underline{59.3} \\
     FairMOT  \cite{zhang2021fairmot} & No &	73.7 & 72.3 & \underline{59.3}\\
    \hdashline
    SimpleReID* \cite{karthik2020simple} & Yes & 61.7 & 58.1 & 46.9 \\
  SimpleReID \cite{karthik2020simple} & Yes & 69.0 & 60.7 & 50.4\\
    
    UTrack \cite{liu2022online} & Yes & 71.8 & 70.3 & \textbf{58.4} \\
   UCSL (ours) & Yes & \textbf{73.0}  &	\textbf{70.4} & \textbf{58.4}	\\

    \hline \multicolumn{5}{c}{MOT15} \\ \hline
     EAMTT \cite{sanchez2016online} & No & 53.0 &	54.0 & 42.5 \\
    TubeTK \cite{pang2020tubetk}  & No & 58.4	& 53.1 & 42.7  \\
    RAR15	\cite{fang2018recurrent} & No & 56.5 & 61.3 & 46.0  \\
    MTrack \cite{yu2022towards} & No & \underline{58.9} & \underline{62.1} & \underline{47.9} \\
    FairMOT \cite{zhang2021fairmot} & No & 55.0 & 60.2 & 45.9 \\
    \hdashline
    UCSL (ours)  & Yes & \textbf{59.1} &59.2 & 46.3  \\ \hline
   \multicolumn{5}{c}{MOT20}  \\
    \hline 
     TransCenter \cite{xu2021transcenter}  & No & 58.5 & 49.6 & 54.1\\
   MTrack \cite{yu2022towards} & No & \underline{63.5} & \underline{69.2} & \underline{55.3} \\
    FairMOT \cite{zhang2021fairmot}& No & 55.7 & 64.6 & 52.5 \\
    \hdashline
     SimpleReID* \cite{karthik2020simple} & Yes &	53.6 & 50.6 & 41.7  \\
    SimpleReID \cite{karthik2020simple} & Yes & 61.8 & 54.8 & 45.5 \\
     
    UCSL (ours) & Yes & \textbf{62.4} & \textbf{63.0} & \textbf{52.3}  \\
     \hline
    \end{tabular}
\end{center}
\caption{Performance on MOT17, MOT15 and MOT20 test sets. ``Unsup'' means unsupervised training. ``*'' denotes using public detections. Bold and underline indicate unsupervised and supervised best metrics, respectively.}
  \label{tab:final results}
\end{table}

\textbf{Comparison on MOT17.}
In this part, we compare our method with some other supervised and unsupervised methods on MOT17. In general, the performance of supervised methods is more advantageous purely in terms of metrics. As an unsupervised approach, we expect it to be as close as possible to state-of-the-art results. 
As shown in Table \ref{tab:final results}, we list some popular methods of joint detection and tracking or embeddings, and our method achieves considerable results, especially on IDF1 and HOTA. 
As the results provided by SimpleReID \cite{karthik2020simple} are based on public detections, for a fairer comparison, we use the detection results of the same detector, i.e., CenterNet \cite{zhou2019objects}, to obtain the corresponding private detection-based results of simpleReID \cite{karthik2020simple}.
Since UTrack \cite{liu2022online} is not tested on MOT17 test set, we replace it with our designed UCSL and conduct experiments under the same hardware conditions on MOT17. The results are shown in the tenth result row of Table \ref{tab:final results}. Based on the same FairMOT+CycAs model, although UTrack \cite{liu2022online} and UCSL are very close on IDF1 and HOTA, our model improves 1.2 in terms of MOTA. Our model outperforms UTrack \cite{liu2022online} in terms of ReID feature extraction with the same detection branch. We notice that IDS is not better compared to other methods, which may attribute to that UCSL tracks more trajectories and has a higher recall.

\textbf{Performance on Other Datasets.}
In addition to MOT17, we also conduct experiments on MOT15 and MOT20, as shown in Table \ref{tab:final results}. Since FairMOT \cite{zhang2021fairmot} uses additional MIX datasets for training besides the CrowdHuman dataset, we train and test this method under the same conditions for a fair comparison.
On MOT15, the performance of our unsupervised UCSL is metrically stronger than the supervised methods on MOTA, and achieves comparable overall performance on other metrics. 

MOT20 is more complex than the scenarios in MOT15 relatively and has a larger amount of data, so the results of MOT20 are improved over those on MOT15. Our model outperforms the unsupervised SimpleReID \cite{karthik2020simple} largely, especially on IDF1 and HOTA. Compared with supervised methods, the results show that our method is already comparable to them.

\begin{table}[t]
\begin{center}
  
    \begin{tabular}{c|c|c|c}
     
    \hline  Method & MOTA $\uparrow$  & IDF1 $\uparrow$ & HOTA $\uparrow$\\
    \hline
     JDE(yolov5s) \cite{wang2020towards}  & 70.2 & 66.6 & 54.1 \\ 
     FairMOT \cite{zhang2021fairmot} & 73.7 & 72.3 & 59.3 \\ 
     CSTrack \cite{liang2022rethinking} & 74.9 & 72.6 & 59.3 \\
     \hline \hline
      JDE(yolov5s) + Ours & 69.6 & 68.0 & 55.7 \\
     FairMOT + Ours & 73.0 & 70.4 & 58.4 \\ \hline
    \end{tabular}
\end{center}
 \caption{ Mehtods on MOT17 under the same paradigm, JDE (joint detection and embeddings). ``yolov5s'' denotes detection branch baseline. The upper and lower parts are supervised and unsupervised methods, respectively. }

  \label{tab:JDE}
\end{table}

\begin{table}[t]
\begin{center}
  \setlength\tabcolsep{3pt}
    \begin{tabular}{c|c|c|c}
     
    \hline Method & MOTA$\uparrow$  & IDF1$\uparrow$ & HOTA$\uparrow$\\
    \hline
    YOLOX \cite{ge2021yolox} + BYTE \cite{zhang2022bytetrack} & 78.8 & 77.0  & 62.7  \\
    CenterNet \cite{zhou2019objects} + BYTE \cite{zhang2022bytetrack}  & 73.1 & 70.0 & 58.9 \\ 
    CenterNet \cite{zhou2019objects} + UCSL(ours) & 73.0 & 70.4 & 58.4 \\ \hline
   
    \end{tabular}
\end{center}
 \caption{Comparison with ByteTrack \cite{zhang2022bytetrack} on MOT17. For a more intuitive comparison, we use YOLOX + BYTE to represent ByteTrack directly.}
  \label{tab:byte}
\end{table}

\textbf{Performance under JDE paradigm.}
Our method is based on the JDE paradigm, considering FairMOT \cite{zhang2021fairmot} as the baseline by default. We show the results of classical and our methods under the same paradigm, as shown in Table \ref{tab:JDE}. Since the JDE \cite{wang2020towards} does not provide results on the MOT17 test set, we retest them under the same conditions. Due to the same paradigm, our approach also can be applied in other JDE-based methods, \textit{e.g}., JDE \cite{wang2020towards}.

\textbf{Comparison with TBD.}
Due to the contradiction between detection and ReID, compared with JDE, indeed, TBD (tracking-by-detection) paradigm could achieve a higher performance limit. But joint training methods output detections and embeddings simultaneously, balancing the accuracy and speed. So under JDE paradigm, we focus on exploring the impact of the unsupervised approach on it, rather than aiming for the state-of-the-art performance. 
 To compare our method with TBD, we consider ByteTrack \cite{zhang2022bytetrack} as the representative for advanced TBD methods, First, it should be noticed that ByteTrack \cite{zhang2022bytetrack} uses trajectories interpolation on MOT17 dataset, which turns it into an offline approach. So we test ByteTrack \cite{zhang2022bytetrack} without interpolation on MOT17, as shown in the first result row of Table \ref{tab:byte}. In our approach, the detection branch uses CenterNet \cite{zhou2019objects} by default, so the comparison between the second and third result lines of Table \ref{tab:byte} demonstrates that the performance impact of our unsupervised approach is comparable to that of BYTE \cite{zhang2022bytetrack} with the same detector. 

 \begin{table*}[t]
\begin{center}
    \begin{tabular}{c| c|c|c|c|c|c|c|c|c|c|c}
    \hline  \multicolumn{2}{c|}{$L_{sc}$} &\multirow{2}{*}{$L_{cc}$} &\multirow{2}{*}{$L_{ac}$}  & \multirow{2}{*}{IDF1 $\uparrow$} & \multirow{2}{*}{HOTA $\uparrow$} &\multirow{2}{*}{MOTA $\uparrow$} & \multirow{2}{*}{MT $\uparrow$} & \multirow{2}{*}{ML $\downarrow$} & \multirow{2}{*}{FP $\downarrow$} &  \multirow{2}{*}{FN $\downarrow$} & \multirow{2}{*}{IDS $\downarrow$}  \\
    \cline{1-2}   $L_{dsc}$ & $L_{isc}$  & &  & & & & & & & & \\
    \hline & CycAs \cite{wang2020cycas} & &  & 59.1 & 49.6 & 69.5 & 38.7\% & 19.0\% & 31309 & 132816 & 7839 \\
     \checkmark &  & &  & 61.5 & 50.8 & 69.9 & 39.3\% & 19.8\% & \textbf{29475} & 132954 & 7314 \\
    & \checkmark   & &  & 66.6 & 54.9 & 69.8 & 38.6\% & 18.1\% & 31341 & 133173 & 5997 \\
     \checkmark & \checkmark & &  & 67.2 & 55.0 & 69.4& 38.6\% & 18.2\% & 32502 & 134808 & 5544 \\
     \checkmark & \checkmark & \checkmark &   & \textbf{68.4} & \textbf{55.6} & 69.8 & 39.6\% & 19.2\% & 32619 & 132228 & 5595 \\
     \checkmark & \checkmark & \checkmark & \checkmark   & 68.2 & 55.5 & \textbf{70.5} & \textbf{40.8\%} & \textbf{16.2\%} & 40569 & \textbf{125004} & \textbf{5208} \\
    \hline
    \end{tabular}
\end{center}
\caption{Performance with different losses on MOT17 test set. 
``CycAs'' represents utilizing original loss function in CycAs \cite{wang2020cycas}.
$L_{sc}$ represents self-contrast loss, where $L_{dsc}$ and $L_{isc}$ represents direct and indirect self-contrast loss, respectively. $L_{cc}$ represents cross-contrast loss, $L_{ac}$ represents ambiguity contrast loss. }
  \label{tab:loss}
\end{table*}

\begin{table}[t]
\begin{center}
    \begin{tabular}{c|c|c|c|c}
    \hline Interval & MOTA $\uparrow$ & IDF1 $\uparrow$ & MT $\uparrow$ & ML $\downarrow$  \\
    \hline 7 & 68.5 & 66.5  & 38.2\% & 20.7\%  \\
    3 & 68.7 & 67.7  & 37.8\% & 20.6\%  \\
    1	& \textbf{70.5} & \textbf{68.2}  & \textbf{40.8\%} & \textbf{16.2\%}  \\
    \hline
    \end{tabular}
\end{center}
\caption{Comparison of different input frame intervals. Based on the current frame, three consecutive frames are taken as input according to the number of frame intervals listed in the table. For the current frame $t$, for example, when the interval is 1, the inputs are frames $t$, $t-1$ and $t-2$.}
  \label{tab:input_frame}
\end{table}

\begin{table}[t]
\begin{center}
    \begin{tabular}{c|c|c|c|c}
    \hline ReID Dim & MOTA $\uparrow$ & IDF1 $\uparrow$ & MT $\uparrow$ & ML $\downarrow$  \\
    \hline 64 & 69.8 & \textbf{68.9} & 38.2\% & 20.2\% \\
    128 & 70.5 & 68.2  & \textbf{40.8\%} & \textbf{16.2\%} \\
    256 &  \textbf{70.7} & 68.4 & 37.7\% & 20.6\% \\
    \hline
    \end{tabular}
\end{center}
\caption{Comparison of different output ReID dimensions.}
  \label{tab:reid_dimension}
\end{table}

\subsection{Ablation Studies}
We conduct ablation experiments on the MOT17 test set, in which we test all contrast losses mentioned above as well as some settings about input frame interval and output ReID dimension.

\textbf{Baseline.} 
We are inspired by the method CycAs \cite{wang2020cycas}. As shown in the first row in Table \ref{tab:loss}, only the triple loss of the original CycAs \cite{wang2020cycas} is used for modeling, aiming to make the probability of the object matching back to itself reach a credible level to ensure cycle consistency. In this method, IDF1 and HOTA are 59.1 and 49.6, respectively. 

\textbf{Self-Contrast Loss.} 
We use both the direct and indirect self-contrast losses to construct the model to extract ReID embeddings better. 
In the direct and indirect self-contrast subparts, we both use the intra-frame cross-entropy loss to construct the loss function, 
bringing the same objects closer together in the feature space and different targets further apart in the feature space. 
As seen from the second, third and fourth rows of Table \ref{tab:loss}, both direct and indirect self-contrast learning have little effect on MOTA, while have significantly improved the IDF1 and HOTA metrics and reduced IDS, demonstrating that our self-contrast similarity learning extracts more discriminative ReID embeddings.

\textbf{Cross-Contrast Loss.}  
We use cross- and consecutive-frames matching for cross-contrast similarity learning, with the aim of reducing the effect caused by mutual occlusion between objects. As can be seen from the fifth row in Table \ref{tab:loss}, on basis of self-contrast similarity learning, the cross-contrast improves IDF1 and HOTA metrics to 68.4 and 55.6 respectively, and there are also different degrees of improvement on other metrics. 

\textbf{Ambiguity Contrast Loss.}
In order to consider both the occluded, disappearing and emerging objects in MOT, we use ambiguity contrast to match these ambiguous objects again. From the last row of Table \ref{tab:loss}, one can see that after adding the ambiguity contrast based on the above two losses, the result has a more obvious improvement mainly in MOTA, MT and IDS, indicating that the method does have a positive effect on maintaining the object’s trajectory.

\textbf{Input Frame Interval.} 
In our model, the default input is three consecutive frames. To show its superiority, we set different input intervals to train and test the corresponding model on MOT17, as shown in Table \ref{tab:input_frame}. Generally speaking, occlusion will last for a long time, but we find the larger the frame interval the weaker the performance, which may be surprising but explainable. 
Large interval is more suitable for supervised settings, where objects between any frames can be well matched with annotated ID labels. However, long intervals may cause drastic object changes without ID labels, making matching hard and errors accumulated. 
In addition, during training, there is an intersection between each input frame group. So, long-term temporal relation is taken into consideration just in an implicit manner. 

\textbf{Output ReID Dimension.} 
In Table \ref{tab:reid_dimension}, we compare three different ReID embedding dimensions. As we can see, compared to the 64-dimension ReID embeddings, the 128-dimension performs better in terms of MOTA and MT metrics. The 256-dimension features have a similar improvement effect on the MOTA and IDF1 as the 128-dimension but consume more space and slow down the training and inference speed. For all these reasons, we choose the 128-dimension as the output dimension of the ReID branch.

\section{Conclusions}
We propose a simple but effective unsupervised method based on Contrastive Similarity Learning (UCSL). Specifically, we construct three learning types: self-contrast, cross-contrast and ambiguity contrast learning. Combining these sub-modules, the network is able to learn discriminative features consistently and reliably, and handle with occluded, lost and emerging objects simultaneously. Our unsupervised method outperforms existing unsupervised methods, and even surpasses some advanced supervised methods.

{\small
\bibliographystyle{ieee_fullname}
\bibliography{egbib}
}

\end{document}